# Reasoning About Beliefs and Actions Under Computational Resource Constraints

Eric J. Horvitz

Medical Computer Science Group
Knowledge Systems Laboratory
Stanford University
Stanford, California 94305

## Abstract

Although many investigators affirm a desire to build reasoning systems that behave consistently with the axiomatic basis defined by probability theory and utility theory, limited resources for engineering and computation can make a complete normative analysis impossible. We attempt to move discussion beyond the debate over the *scope* of the problems that can be handled effectively to cases where it is clear that there is insufficient computational or engineering resource to perform an analysis deemed to be complete. Under these conditions, we stress the importance of considering the expected costs and benefits of applying alternative approximation procedures and heuristics for computation and knowledge-acquisition. We discuss how knowledge about the structure of user utility can be used to control value tradeoffs for tailoring inference to alternative contexts. We finally address the notion of real-time rationality, focusing on the application of knowledge about the expected timewise-refinement abilities of reasoning strategies to balance the benefits of additional computation with the costs of acting with a partial result.

## I. Introduction

Enthusiasm about the use of computation for decision support and automated control within high-stakes domains like medicine has stimulated interest in the construction of systems that behave consistently with a coherent theory of rational beliefs and actions. A number of investigators interested in the automation of uncertain reasoning have converged on the theoretical adequacy of the decision-theoretic basis for rational action. [10, 2, 13]. Recent discussions about computational approaches to reasoning with uncertainty have focused on the degree to which probability and utility theory can handle inference problems of realistic complexity. Investigators have answered criticism about the inadequate expressiveness of probability theory by pointing out that the normative basis focuses only on the consistent inference of belief and value, not on the formulation of the problems [14]. Others have shown that probability theory and utility theory are logically equivalent to the satisfaction of a small set of intuitive properties [28, 14]. Still others have responded to complaints of

---

This work was supported by NASA-Ames under Grant NCC-220-51 and the National Library of Medicine under Grant RO1LM04529. Computing facilities were provided by the SUMEX-AIM Resource under NIH Grant RR-00785.

intractability by demonstrating techniques that can solve relatively complex real-world problems [23, 12].

In this paper, we move beyond discussions of the degree to which the theories of probability and utility are able to solve real-world problems. We focus on situations where it is clear that insufficient resources prohibit the use of the normative basis for a complete analysis. That is, we are interested in studying cases where normative reasoning is clearly inadequate because of pressing resource limitations. We are concerned with rational strategies for handling such resource breaking points. We have been examining resource constraints at the knowledge assessment, computation, and explanation phases of automated reasoning systems.

We will focus here primarily on the example of real-time decision making. Resource constraint issues can be especially salient in the context of real-time requirements. In the real world, delaying an action is often costly. Thus, computation about belief and action often incurs inference-related costs. The time required by a reasoning system for inference varies depending on the complexity of the problem at hand. Likewise, the costs associated with delayed action vary depending on the stakes and urgency of the decision context. The real-time problem is additionally complicated by the existence of uncertainty in the cost functions associated with delayed action. We are searching for uncertain reasoning strategies that can respond flexibly to wide variations in the availability of resources. The intent of our research is to develop coherent approaches to generating and selecting the most promising strategy for particular problem-solving challenges.

## II. Components of Uncertain Reasoning

We have found it useful to decompose uncertain reasoning into three components: problem formulation, belief entailment, and decision making. *Problem formulation* is the task of modeling or constructing the reasoning problem. This task often involves enumeration of all hypotheses and dependencies among hypotheses. There are no formal theories for problem formulation; in many reasoning system projects, engineers charge domain experts with the task of enumerating all relevant propositions. *Belief entailment* or *inference* refers to the process of updating measures of truth assigned to alternative hypotheses as new evidence is uncovered. In most schemes, the degree of truth or *belief* in the presence of a hypothesis can range continuously between complete truth and complete falsity. Finally, *decision making* is the process of selecting the best action to take. A decision or action is an irrevocable allocation of valuable resource.

The classical decision-theoretic basis defines rational beliefs and actions with the axioms of probability theory and utility theory. Probability theory dictates that the assignment and entailment of beliefs in the truth of propositions should be consistent with a set of axioms. The logical equivalence of these axioms with a small set of intuitive properties desired in a measure of belief has been demonstrated [6, 14]. *Utility theory* [29] dictates the consistent assignment and updating of the *value* of alternative actions given the value of alternative outcomes and the degrees of belief in the outcomes. Measures of value consistent with the axioms of utility theory are called *utilities*. Von Neumann and Morgenstern, the authors of utility theory, proved that agents making decisions consistent with the axioms of utility would behave as if they associate utility values with alternative outcomes and act to maximize their expected utility [29].

The application of probability theory for belief assignment and utility theory for decision making defines a *normative basis* for reasoning under uncertainty. The term *normative* refers to the notion that probability theory and utility theory have been accepted in several disciplines as a consistent axiomatic basis for inference that is considered optimal. That is, for many, the normative framework defines a *rational* theory for belief and action.

## III. On the Limited Scope of the Normative Basis

Artificial intelligence research has highlighted the problems that lurk beyond the axiomatic framework defined by probability and utility theory. The real-world problems examined by machine intelligence investigators are often more complex than problems previously tackled with decision theory. In applying the normative basis to many real-world problems, the limited domain of discourse of the theory becomes apparent.

It is clear that significant aspects of problem modeling and inference in the real world are absent from the language and axioms of the normative basis. The normative theory's sole focus on the consistent assignment and inference of measures of belief and preference is dwarfed by the complex task of constructing and solving the uncertainty problem. For example, the axioms have nothing to say about the modeling process. They do not address issues surrounding the most appropriate propositions to represent, the level of abstraction to select, nor the degree of completeness or detail of interdependencies to represent.

The normative basis also does not address issues surrounding the most appropriate inference technique for reasoning problems under specific computational resource constraints. The classical notion of normative rationality implicitly assumes sufficient computational resources for reasoning about an optimal action; the basis does not explicitly address issues surrounding the value of alternative approaches to incomplete inference in reasoning systems that might be dominated by varying limitations in computational or engineering resources.

There is much research to be done on the reformulation of problems and inference strategies deemed optimal in a world with infinite resources to perform in resource-limited environments. In this regard, we see promise in the development of techniques for examining alternative models and inference strategies as the *objects* of design-time and real-time metalevel analysis. This task involves determining, in a tractable fashion, the *most promising* expenditure of engineering or computational resource. Our research has highlighted the notion that a system with the ability to reason under uncertainty on complex real-world problems often requires extensive knowledge about the domain at hand as well as knowledge about the expected behavior of alternative inference strategies.

## IV. The Complexity of Rationality

Let us pause briefly to consider the complexity of normative rationality. Recent research has focused on the computational complexity of probabilistic reasoning. The research has been based upon analyses of uncertain-reasoning problems represented with graphs. The most popular representation uses directed graphs to explicitly represent conditional dependencies and independencies among beliefs in propositions. [22, 4, 19] Many researchers have ascribed a common semantics to the directed graphs. A common term for the representation is *belief networks*.

In a belief network, an arc between a node representing proposition A and one representing proposition B expresses knowledge that the probability distribution over the values of B depend on the specific values of proposition A. If there is no arc from A to B, the probability distribution for B is not directly dependent on the values of A. Less expressive representations commonly employed in artificial intelligence research have not allowed specific independencies

to be represented efficiently [11].[2]

Belief networks are special cases of more general graphical representations that allow actions and the value of alternative outcomes to be represented in addition to beliefs [16, 23]. These graphs have been called *influence diagrams* and *decision networks*. An example of a simple decision network for medical diagnosis is shown in Figure 1. Note that the observed symptoms ($T_r$) are dependent on the disease present, and that the value (V) of the decision to assume a specific diagnosis depends on the disease assumed (Dx) and the actual disease present (D). The possibility of doing additional testing is represented by decision T.

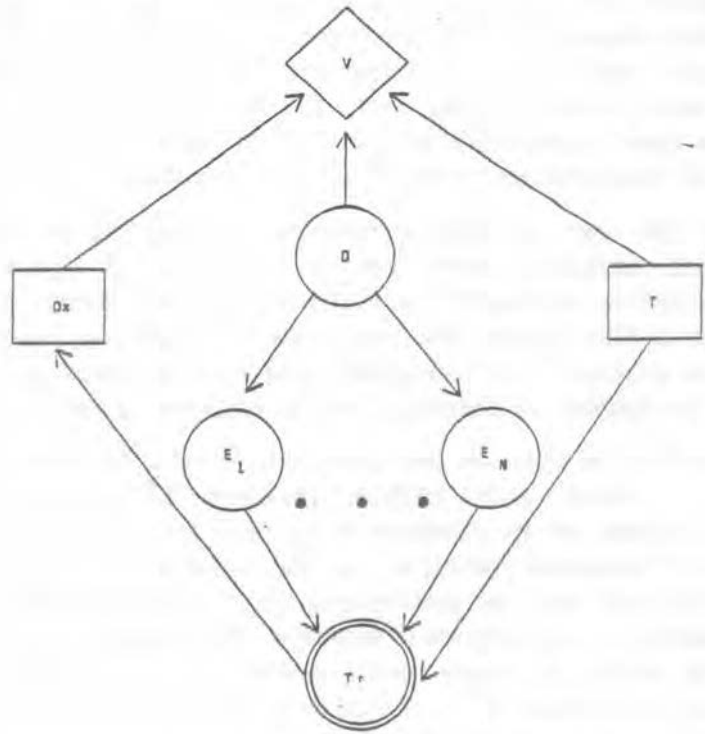

Figure 1: A decision network for diagnosis.

Although the directed-graph representations allow the expression of inference problems that can be solved efficiently, many topologies have resisted tractable algorithmic solution. An example of a difficult problem category is called the *multiply-connected network* [19]. Such inference problems belong to a class of difficult problems which have been proven to have NP-hard time complexity in the worst case [5]. Problems in complex areas like medicine often require representation with multiply-connected networks. Thus, rational beliefs and actions may frequently demand intractable computation.

It is clear that many uncertain reasoning problems require more computation time than may be available before a commitment to action is required. What can be done when the cost of inference becomes intolerable? As a first step, investigators might search for special-case inference techniques designed for the efficient solution of specific problem types (e.g. specific belief network topologies or belief distributions). However, proofs such as the one demonstrating the worst-case intractability of multiply-connected networks, put little hope in the discovery of special methods that will solve some important classes of problems. For many situations, we will need to develop intelligent approximation procedures and heuristics[3] that focus the expenditure of resources on the most relevant aspects of the uncertain reasoning problem at hand.

---

[2] For example, the popular rule-based representation may encourage researchers to make global assumptions about the absence of dependencies among propositions.

The pressures of complex decision making in real-time force Bayesian theoreticians and engineers to consider alternatives to normative reasoning; under time constraints (or other resource constraint like knowledge-acquisition cost), approximations and more poorly-characterized heuristic techniques can have higher expected value than complete normative reasoning. The delay associated with inference might be so costly that an alternative approximation or heuristic strategy would have a greater expected value, in spite of assured suboptimality or uncertainty in the performance of the strategy. Thus, constraints in resource can transform a non-normative technique into the "preferred choice" of devout Bayesians and convert the strictest formalists into admirers of heuristics.

We have been investigating the problem of reasoning under specified constraints within the PROTON project. A focus of research centers on intelligent control strategies for selecting among alternative problem-formulation and inference strategies. We are studying decision-theoretic approaches to control. We believe that the representation of explicit knowledge about the costs associated with computation such as time-delay will be useful in complex uncertain reasoning problems. Although we hope to discover approximate inference techniques that show clear dominance, we believe that it may often be important to reason about inference tradeoffs under uncertainty at the metalevel.

## V. Inference Under Resource Constraints

Simple normative reasoning systems have been based on a single model constructed as a static basis and acted upon by a single inference strategy. We are interested in techniques for reformulating a basis problem into one that will be of greater value than a complete analysis under computational resource constraints. A reasoning system with knowledge about the behavior of alternative approximation and heuristic strategies and about the costs associated with inference-based delay might provide valuable computation under resource constraints that would render a complete normative analysis to be a worthless or costly enterprise.

We will now raise several issues about strategies that can focus computational attention on the most relevant portions of uncertain reasoning problems. Challenging components of this research include the development of approximation procedures and heuristics that are insensitive to small variations in resource availability, the representation of knowledge about the value structure of the problem, and the development of compiled and real-time control strategies that can recognize problems, understand the problem-solving context, and select the most valuable inference strategy.

### Integrating Knowledge About Inference-Related Costs

It is clear that theoretical models of rationality should include the costs associated with rational inference itself. Previous research has touched on the integration of the costs of reasoning into decision making inference [8, 24, 17]. Within the realm of automated reasoning, representing inference costs can be valuable in the control of inference. A crucial aspect of integrating knowledge about real-world costs, benefits, and tradeoffs into a reasoning system is the acquisition of knowledge about the value of important attributes of computer performance to the users of computer systems.

---

[3] We use *approximation* to refer to a strategy that produces a result with a well-defined margin of error; we use the term heuristic to refer to strategies which have uncertain performance. A strategy may be viewed as *heuristic* in terms of specific aspects of its behavior. According to this perspective, investigation leading to new characterization of a strategy can transform a "heuristic method" into an approximation strategy.

We have found it useful to decompose the value associated with computational inference into several components. We assert that the application of an inference strategy is associated with some net benefit or cost to an agent such as a system user, a robot, or a computational subsystem, relying on computation for decision making. We use the term *comprehensive value* ($V_c$) to refer to the net *expected utility* associated with the application of a computational strategy. We will see that this value is a function of the strategy, of the problem, and of the problem-solving context. We have found it useful, in studying inference tradeoffs under pressing resource limitations, to view the comprehensive value as having two components: the *object-related* value and *inference-related* value.

The *object-related* value ($V_o$) is the expected utility associated with computation-based increases in information about the *objects* of problem solving. For example, the object-related value associated with the use of an expert system for assistance with a complex medical diagnosis problem refers to the costs and benefits associated solely with the change in information about the entities in the medical problem such as alternative treatments, likelihoods of possible outcomes, and costs of recommended tests.

The *inference-related* value ($V_i$) is the expected disutility intrinsically associated with *computation*, such as the cost a physician might attribute to the delay of a decision because of the time required by an expert system to generate a recommendation, or the cost associated with his inability to understand the rationale behind a decision recommendation. We will later describe an example that shows how representing the cost of computation in different contexts can be crucial.

We have decomposed the expected utility of a computational process into two components for presenting issues about inference-related cost. In general, we may have to consider important the dependencies between the object- and inference-related value. We assume the existence of a function F that relates $V_c$ to $V_o$, $V_i$ and additional background information about the problem-specific dependencies that may exist between the two components of value. That is,

$$V_c = F(V_o, V_i, \phi)$$

where $\phi$ captures problem-specific background information about possible dependencies between object- and inference-related value.

Knowledge about costs and benefits of computation can be integrated into the decision network representation. A more comprehensive representation of our simple diagnosis problem is portrayed in Figure 2. Note the new arcs and nodes that capture autoepistemic knowledge about the costs associated with computation, as well as the new decision node reflecting metalevel reasoning costs and metalevel decision making about the form of the object-level problem. The metalevel reasoning problem is to optimize the comprehensive value ($V_c$).

### Assigning utility to multiple attributes of inference

The components of value described above can be ascertained through assessing important attributes of computational performance through a subjective assessment of the value of alternative performance scenarios to a system user or through the construction of a function capturing the relationships among attributes of computational value in important contexts. The value assigned to alternative computational behaviors can often be described by a qualitative or more detailed function that represents the relationships among important components of perceived costs and benefits associated with alternative outcomes. Such value functions assign a single value measure to computation based on the status of an $n$-tuple of attributes. For example, the value associated with the use of a medical expert system in a particular context might be a function of a number of attributes, including speed of computation, accuracy of recommendation, and clarity of explanation. We have been working with expert physicians in the intensive-care and tissue pathology domains to ascertain value models relating measures of utility to multiple attributes of computation.

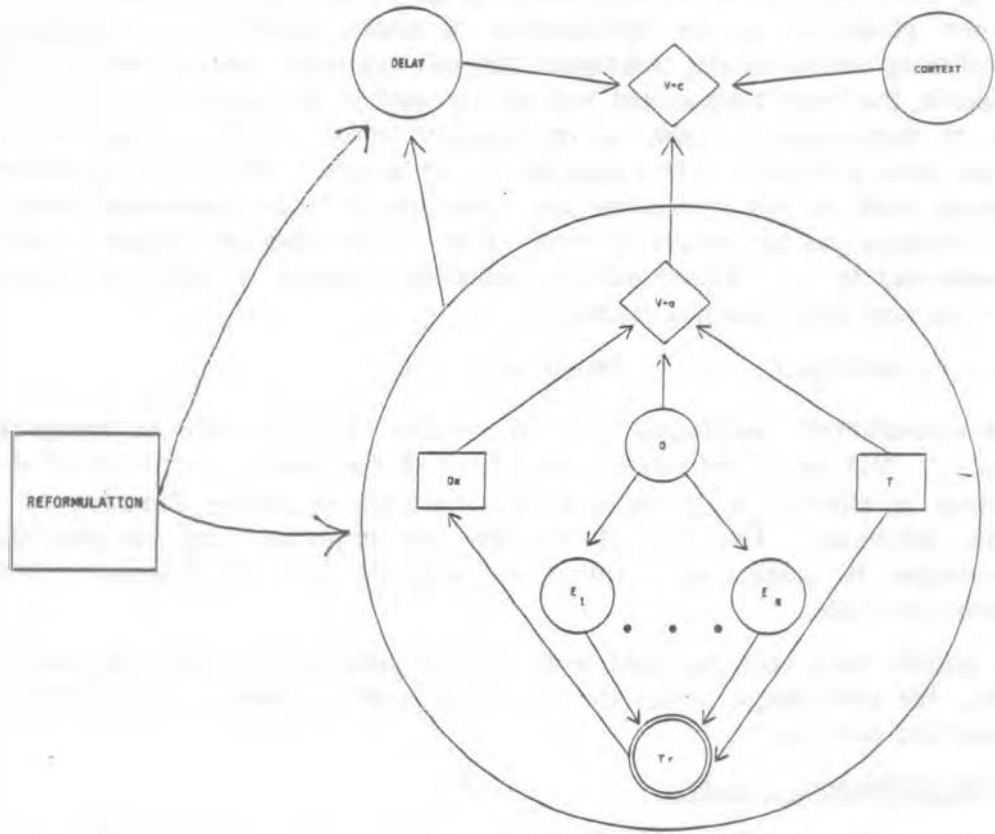

Figure 2: A decision network for a more comprehensive repesentation of the problem of diagnosis.

We are not the first to explore the formal use of utility theory in the control of reasoning. Concurrent research has focused on the usefulness of assigning utilities to alternative strategies in the control of logical reasoning [25, 27]. The research presented here differs from the other work in its focus on representing multiple components of value and on the integration of context-specific knowledge concerning human preferences about computational tradeoffs.

Inference Tradeoffs

Computation in a world of bounded resources often is associated with cost/benefit *tradeoffs*. Working with expert physicians on the development of expert systems has highlighted the importance of developing computational techniques that can explicitly control tradeoffs. With a computational tradeoff, the benefit associated with an increase in the quantity of one or more desired attributes of computational value is intrinsically linked to costs incurred through changes imposed on other attributes. More specifically, we define a tradeoff as a relationship among two attributes, such as the *immediacy* and *precision* of a computational result, each having a positive influence on the perceived total value of computer performance, such that they are each constrained to be a monotonically decreasing function of the other over some relevant range. In the case of our sample tradeoff,

$$\text{PRECISION} = F(\text{IMMEDIACY}), \quad t_o \leq \text{IMMEDIACY} \leq t_n \qquad (1)$$

where $F$ is some monotonically decreasing function over the range bounded by computational time delays $t_o$ and $t_n$. This definition can be generalized to the case where the value assigned to tuples of a subset of relevant attributes is a monotonically decreasing function of tuples composed of other attributes. The tradeoff between the immediacy and the precision or accuracy of a solution is particularly explicit in methods that incrementally refine a computational result with time.

Most reasoning systems have been designed with implicit assumptions about the handling of inference tradeoffs. We have sought to develop tools that enable systems to tailor inference to a range of problems and contexts.

### Toward a Timewise-Refinement Paradigm

Let us now focus on the properties of approximate and heuristic inference that would be useful under varying resource constraints. Classical approaches to normative inference have focused on the determination of point probabilities. In fact, the complexity proof described above is based on the assumption that point probabilities are required. The classical interest in calculating final answers permeates computer science. Complexity theorists have focused almost exclusively on proving results about the time and space resources that must be expended to run algorithms to termination [7, 1, 18]. In the real world, strict limitations and variations on the time available for problem solving suggest that the focus on time complexity for algorithmic termination is limited; analyses centering on how *good* a solution can be found in the time available for computation are of importance.

The major rationale for the focus on the time complexity of algorithmic termination seems to reside in the simplifying notion in algorithms research that a computer-generated result can be assigned only one of two measures of utility: either a solution is found and is of value, or a solution is not found and is therefore valueless. However, it is often possible to enumerate representations and inference techniques that can provide partial solutions of varying *degrees* of value.

An approach to developing techniques for optimizing the value of uncertain reasoning under ranging resource limitations is the development of problem reformulation and inference schemes that allow the generation and efficient manipulation of *partial results*. We are interested in representation and reasoning methods that allow a result to be refined with increasing amounts of computation. In analyzing the timewise-refinement behavior of algorithms, it is crucial to consider knowledge about the *value structure* of partial results. We believe that a metalevel formalization of the costs and benefits, and the cost-benefit tradeoffs, associated with inference in differing contexts will be beneficial in the development of insights about useful approximations and heuristics.

### Describing Resource Limitations

We would like to enumerate properties of inference strategies that can be of value under conditions of incomplete resources. Before we enumerate several desirable properties, we must focus more closely on issues surrounding resource availability.

A *resource* is some costly commodity required for inference; we have been focusing on computational time. We define the minimum amount of computational resource needed to solve what has been deemed a *complete* description of the object-related problem as the *complete resources* ($R_c$). It is clear that all models are incomplete to some extent: we use *complete* to refer to an object-related model perceived to be an adequate representation of a problem by a system user or expert.

The *complete* resource level is a function of the complete problem description at hand. We refer to the complete resources more specifically as $R_c(I,P)$, where I is an inference strategy and P refers to a problem defined by a model and a context. We define the object-related value associated with the application of a normative inference strategy in the context of complete resources as the *optimal object-related value*, written [$V_o$]. We call the resource actually applied to problem solving the *allocated resources* ($R_a$) and the call the ratio of the allocated and complete resources the *resource fraction*, $R_f(I,P)$. $R_f(I,P)$ has been a useful metric for communicating about computation under bounded resources.

We can use the defined notions of resource fraction, comprehensive value, object-related value, and inference-related value to express properties desired of inference strategies applicable in environments dominated by varying resource limitations.

### Desiderata of Bounded-Resource Computation

We are interested in representation and control strategies that can configure knowledge and processing in a manner that make effective use of information about the uncertainty in the amount of computational resource available for computation in alternative contexts. For example, we wish to implement representation and inference methodologies that allow the most relevant updating to occur early on. Also, as many real-world applications may involve reasoning under large variations in the time available for inference, it is desirable to design inference strategies that are insensitive to small ranges in resource fraction.

We now enumerate desiderata desired of computational inference. under resource limitations. The desiderata address the usefulness of a graceful response to diminishing resource levels. Desired properties of bounded-resource computation are as follows:

1. <u>Value dominance.</u> We seek problem-solving strategies with value-dominant intervals over available quantities of resource. We define *value-dominant* intervals as ranges of resource fraction over which the gain in the comprehensive value of computation is a monotonic-increasing function of resource.

2. <u>Value continuity.</u> We desire the comprehensive value, the object-related value, and inference-related value to be continuous functions of the resource fraction as it ranges from zero to one. That is,

$$\lim_{R_f \to R_f'} V_c(I, P, R_f) = V_c(I, P, R_f')$$

where $V_c(I, P, R_f)$ is the expected value of computation associated with applying inference strategy I to problem P, with resource fraction $R_f$. We refer to the *continuous* decrease of object-related value with decreasing allocation of resource over value-dominant resource ranges as *graceful degradation*.

Although continuity in the components of value is desirable in providing a continuum of options under pressing resource constraints, it often is difficult to generate such continuity within the discretized realm of computation. Thus, *value continuity* may be generalized to *bounded discontinuity*, where a desired upper bound on an $\epsilon$ change in $V_c$ is specified for some $\delta$ change in $R_f$ over ranges of resource. We have found it useful to represent knowledge about an inference strategy's behavior in terms of constraints on $\delta$ and $\epsilon$. The statement of such constraints or of a probability distribution over such constraints for particular contexts can be used as a partial characterization of heuristic strategies for important aspects of performance.

3. Bounded optimality. The third desideratum is a meta-analytic property describing inference choice. We desire a reasoning system to select an inference strategy or sequence of strategies from the set of strategies available to it such that the comprehensive value of computation is a maximum given a problem, resource fraction, and metalevel resource allocation. That is, a strategy or tuple of strategies $(I_1..I_n)$ should be selected from the set $\{I\}_\epsilon$ of all available strategies that maximizes the expected utility value:

$$(I_1..I_n) : \text{Max} \quad V_c\{I\}_\epsilon$$

A system satisfying the bounded optimality property captures notions of rationality under resource constraints. Such a system attempts to optimize the expected value of its computation regardless of the method lying at the foundations of its inference.

Finally, we note that the *value continuity* and *bounded optimality* properties imply that the object-related value will demonstrate *endpoint convergence* to the optimal object-related value as the resource fraction approaches one. That is, a reasoning system will revert to object-level rationality with complete resources.

$$\lim_{R_f \to 1} V_o(I, P, R_f) = [V_o]$$

## VI. Bounded-Resource Reasoning Strategies

Several classes of approximation methods and heuristics are promising sources of useful strategies for bounded-resource computation. We enumerate several approaches below. Although we group the methods into approximation and heuristic categories, it is clear that analysis of specific instances of the heuristic approaches could lead to crisp approximation procedures.

### Approximation Methods

Bound calculation and propagation. There has been ongoing interest in the calculation of upper and lower *bounds* on point probabilities of interest [4]. Probabilistic bounding techniques determine bounds on probabilities through a logical analysis of constraints acquired from a partial analysis. Such techniques can be configured to focus attention on the most relevant aspects of the uncertainty problem. Bounds become tighter as additional constraints are brought into consideration. Cooper [4] has applied a best-first search algorithm to calculate bounds on hypotheses.

Stochastic simulation. Simulation techniques are approximation strategies that report a mean and variance over the probabilities of interest through a process of weighted random sampling [12, 20]. In many cases, the distribution over the probabilities is approximated by the *binomial* distribution. The variance with which the mean converges with additional computation depends on the topology and the nature of the probabilistic dependencies within the network. Recent work [3] has shown current simulation algorithms to have intolerably slow convergence rates in many realistic cases. Stochastic simulation is nevertheless a promising category of inference for the derivation of useful bounded-resource computation strategies.

**Heuristic Methods**

Completeness modulation. Completeness-modulation strategies focus on techniques for reasoning about aspects of the uncertain reasoning model to include in an analysis. Completeness modulation may be used to simplify the topology of a belief network through deleting classes of dependencies. In one form of completeness modulation, arcs in the graph are prioritized by heuristic measures of context-dependent "importance" that capture the benefits of including the dependencies in alternative contexts. Such heuristic measures may be encoded during knowledge acquisition. The measures allow a reasoning system to dynamically construct a model that will be subjected to some inference procedure (e.g. bounding, simulation, complete normative analysis). Under time constraints, a completeness modulation approach can allow components of the problem viewed as most important to be included early on in an analysis. We have worked with experts to acquire measures of *importance* on probabilistic dependency a medical domain.[4]

Abstraction modulation. In many cases, it may be more useful to do normative inference on a model that is deemed to be complete at a particular level of abstraction than to do an approximate or heuristic analysis of a model that is too large to be analyzed under specific resource constraints. It may be prove useful to store several belief network representations, each containing propositions at different levels of abstraction. In many domains, models at higher levels of abstraction are more tractable. As the time available for computation decreases, network modules of increasing abstraction could be employed.

Imposition of global independence. A long-standing heuristic in reasoning under uncertainty involves the assumption (or the imposition) of large-scale independence among propositions considered by a system. Such an assumption greatly reduces the resources required for knowledge assessment and computation. Global conditional independence assumptions have been made in many reasoning systems that have been deemed to perform adequately (e.g. the MYCIN certainty-factor model [10, 13] and innumerable early "tabular Bayes" diagnostic programs [9, 26]). While it is easy to construct examples where the assumption of conditional independence induces severe pathology, the actual costs and benefits of assuming conditional independence among evidence in many real-world problems have not been determined.

Local reformulation. Local reformulation refers to the modification of specific troublesome topologies in a belief network. Approximation methods and heuristics focused on the microstructure of belief networks will undoubtedly be useful in the tractable solution of large uncertainty reasoning problems. Such strategies might be best applied at knowledge encoding time. An example of a potentially-useful local reformulation is the use of tractable *prototypical dependency* structures such as the noisy-OR structure. [21]. The benefits of using such structures for knowledge acquisition and inference could warrant the use of tractable prototypical dependencies in situations where they are clearly only an approximation of more complex dependencies.

---

[4]The use of importance measures may also be useful in directing the allocation of resources during knowledge assessment.

<u>Default reasoning and compilation.</u> Under severe time pressure, general default beliefs and policies may have more expected value than a computed result. Indeed, in some application areas, it may be useful to focus a reasoning system's scope of expectation through the compilation and efficient indexing of computed advice for actions of great importance, high-frequency, or that are frequently needed in time-critical situations. The relative worth of storing heuristic default knowledge or compiled policies depends on a number of factors, including the tractability of available inference strategies, the nature of the available resource fraction, and the complexity of expected outcomes in the application area. Decisions on whether to compute or to store recommendations may also be quite sensitive to the specific costs of computer memory and knowledge assessment. Careful consideration of the value structure of components of computation in real-time and in system-engineering settings can help to elucidate specific cases of such tradeoffs.

### The Intelligent Control of Uncertain Inference

Techniques for different categories of inference mentioned above could be combined to generate useful classes of bounded-resource strategies. Such classes might be constructed and tailored to the categories of time constraints within a particular application area during the engineering of a system. For example, multiple representations of a problem, each tailored to maximize the value of computation in contexts with differing temporal cost-functions might be stored in conjunction with simple application rules.

Attempting to satisfy the *bounded-optimality* property mentioned above may involve intelligent *real-time* metalevel reasoning, requiring the development of techniques for efficient real-time problem recognition, problem decomposition, strategy selection, and strategy monitoring. Complex real-time metalevel reasoning will also require management of the costs and benefits of metalevel inference.[5] We are currently studying the usefulness of metalevel reasoners with access to a several base-level strategies and with rich control knowledge about the value of the strategies in different problem contexts.

## VII. Metalevel Reasoning About the Time-Precision Tradeoff

We will now exercise several of the concepts presented with an example that is representative of ongoing research. We focus on the use of knowledge about multiple components of value at the metalevel to tailor inference to the appropriate context. The example reflects ongoing work on the PROTON system [15] for reasoning about inference tradeoffs. Although the results can be derived formally, we will describe the sample problem with a set of qualitative curves for clarity. The curves capture important functional relationships among components of computational value in alternative contexts.

Consider an inference problem from one of our application areas: An automated control system is faced with a rapidly evolving set of respiratory symptoms in a patient in an intensive-care unit. Assume that our system's action depends on $P(C \mid E)$--the probability of a condition C given the observed symptoms E. In particular, this probability is important in deciding whether or not the systems will respond with a costly treatment for condition C.

What kinds of strategies might our autonomous pulmonary decision making system employ to respond rationally under pressing time constraints? Assume that the system has a base model deemed by a human expert during knowledge acquisition as an *adequately complete model* of aspects of the world that compose the system's domain of applicability.

---

[5] It is clear that empirically- or heuristically-determined limits on metalevel effort will have to be imposed; if not, there is a problem with infinite analytical regress.

Figure 3(a) demonstrates the knowledge that the medical decision system may have about the expected rate of computational refinement of the precision of the requested probability for stochastic-simulation strategy, E-1, given this type of problem.

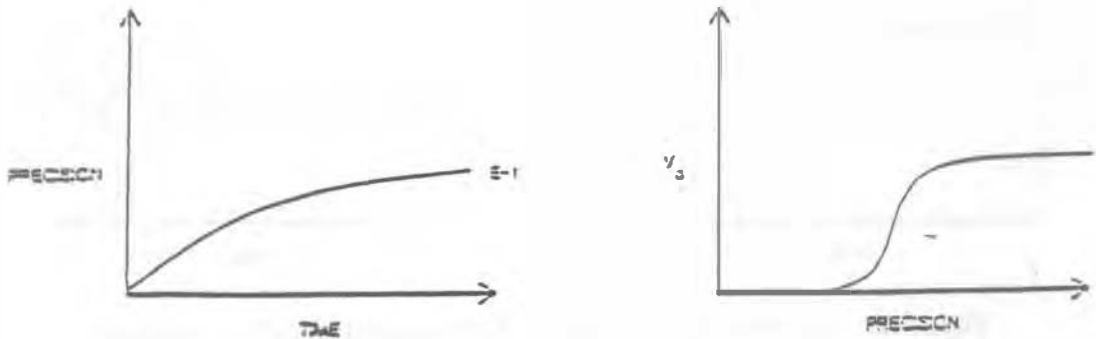

Figure 3: (a) Precision over time. (b) Object-related value.

Let us now introduce computational value considerations. The particular assignment of value to results of increased precision depends on the decision context; the value of an imprecise probability to a user can range greatly depending on the end use of the probabilistic information. A system could be endowed with knowledge about the changes in expected value of perfect information with additional inference.

To encode knowledge about the assignment of object-related value to partial probabilistic results of different precisions, we could work with an expert to assess the utility directly, apply some preenumerated value function, or formally analyze the decision-making context. Let us briefly examine the last option.

Utility theory dictates that the object-related value, $V_o$, in this simple problem, is determined by the probabilities and utilities of four possible outcomes: the patient either has or does not have the condition, and the system will either treat or not treat for the condition.

Simple algebraic manipulation can be used to show that the optimal object-related value of information depends upon the costs associated with treating a person without the condition, the benefits of treating a person with the condition, and the probability of the condition. Thus, changes in the information about the actual probability of the disease can be assigned a measure of value within the decision-theoretic framework.

Let us assume that the expert system has actively acquired information about the context in which the desired probability will be used and has characterized the object-related value of the probability of the condition as a function of the precision of the reported probability. A plausible value function for this situation is shown in Figure 3(b). The function demonstrates that the rate of refinement of the object-related value can vary greatly with increasing precision.

So far, we have examined only object-related value considerations. In the real world, time delay can be quite costly. All the while we have been dwelling on issues regarding the refinement of the object-related value, our patient has been gasping for breath. In this case, it is clear that, for any fixed measure of object-related value, the comprehensive value of the result decreases with the amount of time that a user must wait for it to become available. It is thus important for a medical decision system to have knowledge of the inference-related utility associated with computational inference.

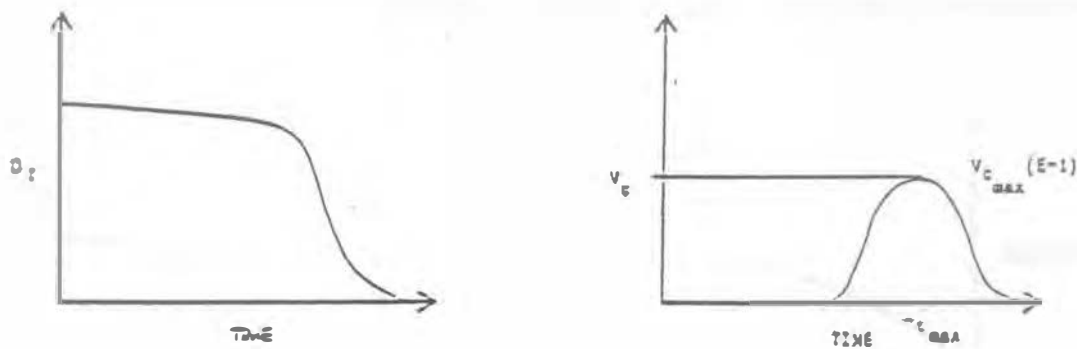

Figure 4: (a) Inference-related cost, (b) Comprehensive value of computation in a decision context.

Let us assume that a physician with extensive knowledge about the realm of possibility in the intensive-care unit had, at an earlier date, represented context-specific knowledge about the rate with which the object-related value should be discounted with the passage of time. That is, it was determined, through utility assessment at the time of knowledge engineering, that the expert physician's preferences about the cost of delay in such a context could be represented as an independent multiplicative discounting factor, $D_t$, ranging in value between one and zero with time.[6] This means that the object-related value is multiplied by the inference-related utility_discounting factor to generate the net value of an answer as time passes. A function demonstrating such a degradation of the object-related value with time is shown in Figure 4(a).

If the information in the three functions are combined, the comprehensive value, $V_c$, of the computational process to a system user as a function of time can be derived. This result is displayed in Figure 4(b).

Notice that the comprehensive value has a global maximum $V_{c_{max}}$ at a particular time, $t_{max}$. This is the period of time the computer system should apply inference scheme E-1 to maximize the value of its reasoning to the patient. Although spending additional time on the problem will further increase the precision, the comprehensive value to the user will begin to decrease. Integrating a consideration of the cost and benefits of computation into an analysis of probabilistic inference makes it clear that the cost of computation can render the solution of the complete problem inappropriate.

<u>Reasoning About Alternative Strategies</u>

So far, we have considered characteristics of the computational value of only one reasoning strategy. Assume that the system's metalevel reasoner has knowledge about the existence of another inference strategy, E-2, based on the modulation of problem completeness. Assume further that the expected precision over time of the more heuristic completeness modulation strategy is represented by the curve portrayed in Figure 5(a). Finally, assume that the system has knowledge that, within this context, the strategy is known to have a higher expected rate of refinement of precision early on, but a lower long-range rate of refinement than that of stochastic simulation.

---

[6] We have considered this factor independent for the simplification of presentation; such a discount rate may depend on the status of the probabilities and outcomes. In this example, we have framed inference-related knowledge acquisition at the level of *classes of criticality* associated with unresolved pathophysiology.

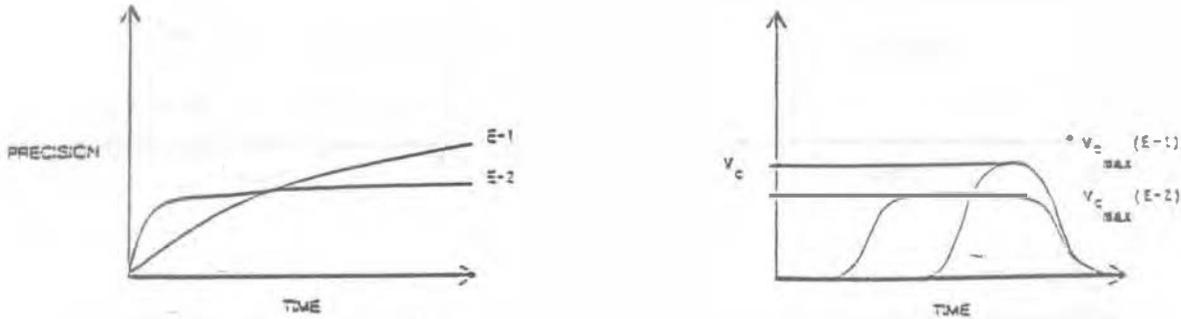

**Figure 5:** (a) Another inference strategy. (b) Comparison of the comprehensive value of the two inference strategies.

If we apply the same object-value and inference-related functions presented previously to the new inference strategy, we can derive a new comprehensive value function. This function is shown in comparison to the previously derived comprehensive value function in Figure 5(b). Notice that

$$V_{c_{max}}(E-1) > V_{c_{max}}(E-2) \tag{2}$$

Equation 2 shows us that a control strategy applying the *bounded-optimality* property would select strategy E-1 given all current knowledge about available probabilistic inference strategies and the decision context at hand.

### Contraction of the Decision Horizon

Now, suppose that the decision context has changed in a way that affects only the inference-related cost function describing the discounting of object-related value with time. In the new context, we have a much sharper discounting of the object-related value with time, as shown in Figure 6(a). Such a decreased decision horizon may be associated with situations requiring rapid response, as might be the case when our patient suddenly begins to show critical signs of poor oxygenation.

If we derive the comprehensive value functions for inference strategies E-1 and E-2 with the new object-related value discounting function, we see a new dominance. Figure 6(b) shows that:

$$V_{c_{max}}(E-2) > V_{c_{max}}(E-1) \tag{3}$$

That is, in contexts of greater extreme time criticality, the value achieved by strategy E-2 will dominate that achieved by E-1 and thus E-2 will be the strategy of choice.

### Defaulting to Default Knowledge

We have focused so far on strategies that can provide partial results through computation. Before concluding, we will move beyond uncertain inference to examine default reasoning.

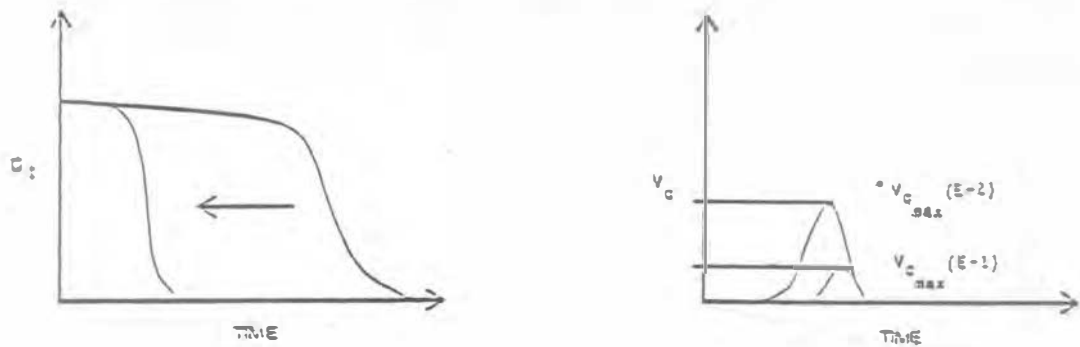

Figure 6: (a) A contraction of the decision horizon. (b) A new dominance in a more time-critical context.

The default strategy E-3 is shown in Figure 7(a). As portrayed in the figure, a default rule for a particular context often can be made available with relatively little computation. Notice that the object-related value of the default strategy within a problem context does not change with time; after being made available, the object-related value of a default strategy is not refined with computation. In this case, we portray the maximum object-related value of the default rule that would "fire" in the context at hand as being a fraction of the object-related value attainable through computation.

A compiled policy with a relatively low object-related value could be the strategy of choice in situations of extreme time criticality. For example, if our patient's blood pressure were suddenly to fall greatly, a theoretically-suboptimal "compiled" default strategy requiring little computation might dominate. We depict graphs reflecting this situation in Figure 7(c) and (d).

We have described the simple example of diagnosis under conditions of pressing time constraints to demonstrate how a reasoning system can apply knowledge about the costs and benefits of alternative inference strategies to optimize the value of computation to a system user. The example demonstrates how classic normative reasoning might be modified to respond to ranging resource constraints.

## VIII. Summary

We have reviewed several issues about decision making under resource constraints. We began the paper with a discussion of the limited scope of the normative basis for reasoning under uncertainty in the real world. We then described the application of knowledge about inference related cost in systems that reason under uncertainty, touching upon the assignment of measures of utility to multiple attributes of computation and the notion of computational tradeoffs. After enumerating desirable properties of bounded-resource inference, we discussed classes of approximation procedures and heuristics that promise to be useful in reasoning under resource constraints. Finally, we described an example that is representative of continuing investigation on the costs and benefits of alternative inference strategies in different settings. We believe that continuing research on the representation and control of uncertain reasoning problems under conditions of varying computational and engineering resources will be crucial for building systems that can act effectively in the real world.

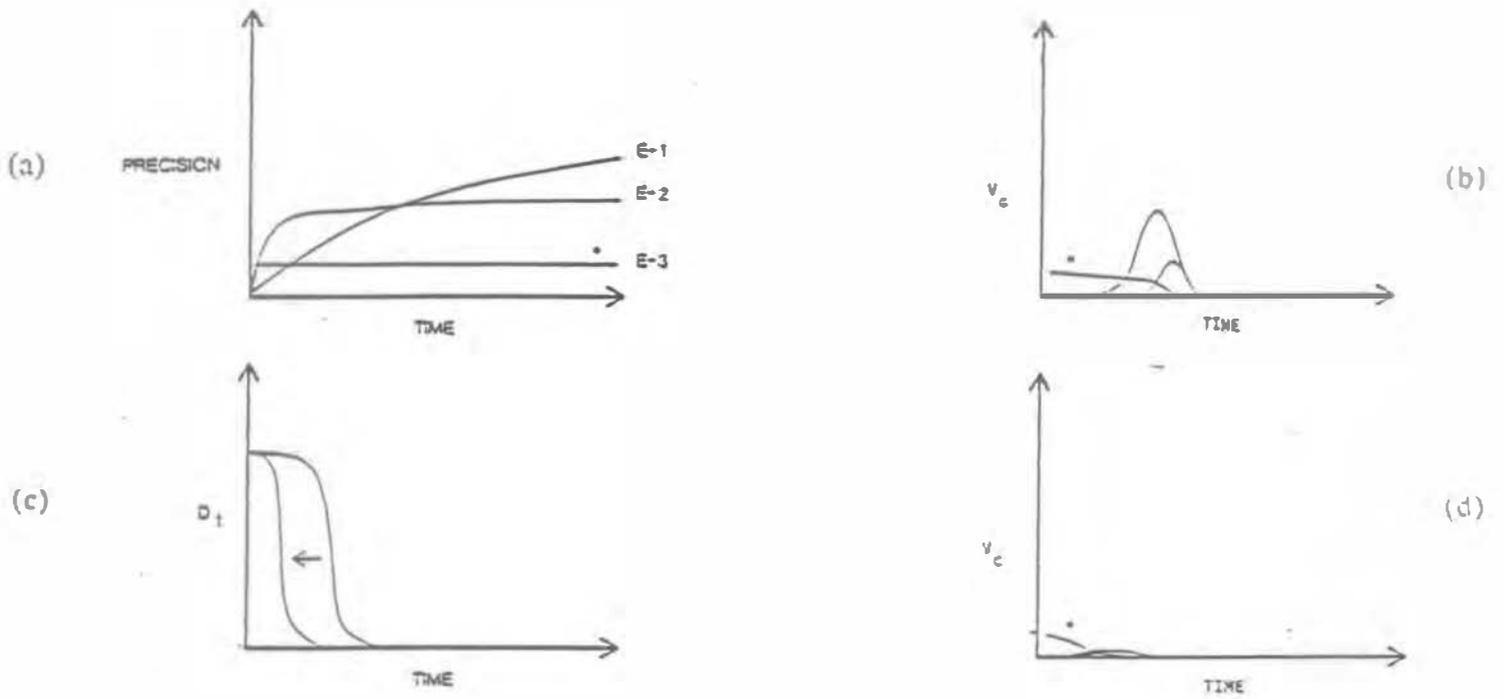

Figure 7: (a) A default reasoning strategy. (b) The comprehensive value of the default strategy. (c) Another shift in the decision horizon. (d) Dominance of the default strategy value under more severely limited computational resource.

## Acknowledgements

I thank Greg Cooper, Bruce Buchanan, George Dantzig, David Heckerman, Ronald Howard, Nils Nilsson and Edward Shortliffe for providing useful feedback on this work.

# References

[1]   Aho, A.V., Hopcroft, J.E., Ullman, J.D. *Data Structures and Algorithms*. Addison-Wesley, Menlo Park, California, 1983.

[2]   Cheeseman, P. In defense of probability. In *Proceedings of the Ninth International Joint Conference on Artificial Intelligence*. IJCAI-85, 1985.

[3]   Chin, H.L. *Stochastic Simulation of Causal Bayesian Models*. Technical Report KSL-87-22, Stanford University, Stanford University Knowledge Systems Laboratory, March, 1987.

[4]   Cooper, G.F. *NESTOR: A Computer-Based Medical Diagnostic Aid that Integrates Causal and Probabilistic Knowledge*. PhD thesis, Medical Information Sciences, Stanford University, 1984. HPP report no. 84-48.

[5]   Cooper, G.F. *Probabilistic Inference Using Belief Networks is NP-hard*. Technical Report, Stanford University, Stanford University Knowledge Systems Laboratory, May, 1987.

[6]   Cox, R. Probability, frequency and reasonable expectation. *American Journal of Physics* 14(1):1-13, 1946.

[7]   Garey, M.R. and Johnson, D.S. *Computers and Intractability: A Guide to the Theory of NP-Completeness*. W.H. Freeman and Company, New York, 1979.

[8]   Good, I.J. Rational Decisions. *J. R. Statist. Soc. B* 14:107-114, 1952.

[9]   Gorry, G. A. and Barnett, G. O. Sequential diagnosis by computer. *Journal of the American Medical Association* 205:849, 1968.

[10]  Heckerman, D.E. Probabilistic interpretations for MYCIN's certainty factors. In Kanal, L.N. and Lemmer J.F. (editors), *Uncertainty in Artificial Intelligence*, pages 167-196. North Holland, New York, 1986.

[11]  Heckerman, D. and Horvitz, E. On the expressiveness of rule-based systems for reasoning with uncertainty. In *Proceedings of the AAAI*, pages 121-126. AAAI, Morgan Kaufman, July, 1987.

[12]  Henrion, M. Propagating uncertainty by logic sampling in Bayes' networks. *Proceedings of the Workshop on Uncertainty in Aritificial Intelligence*, Philadelphia, PA, August 7-10, 1986.

[13]  Horvitz, E. J., and Heckerman, D. E. The inconsistent use of measures of certainty in artificial intelligence research. In *Uncertainty in Artificial Intelligence*, pages 137-151. North Holland, New York, 1986.

[14]  Horvitz, E. J., Heckerman, D. E., Langlotz, C. P. A framework for comparing alternative formalisms for plausible reasoning. In *Proceedings of the AAAI*. AAAI, Morgan Kaufman, Palo Alto, California, August, 1986.

[15]  Horvitz, E.J. *Inference Under Varying Resource Limitations*. Technical Report, Stanford University, 1987. Knowledge Systems Lab Technical Report KSL-87-16, Stanford University, Stanford, California, February, 1987.

[16] Howard, R. A. and Matheson, J. E. *Readings on the Principles and Applications of Decision Analysis*. Strategic Decisions Group, Menlo Park, CA, 1984. (2nd Edition).

[17] March, J.G. Bounded Rationality, Ambiguity, and the Engineering of Choice. *Bell Journal of Economics* , 1978.

[18] Papadimitriou, C.H., and Steiglitz, K. *Combinatorial Optimization: Algorithms and Complexity*. Prentice-Hall, Inc., Englewood Cliffs, New Jersey, 1982.

[19] Pearl, J. On evidential reasoning in a hierarchy of hypotheses. *Artificial Intelligence* 28:9-15, 1986.

[20] Pearl, J. *Evidential Reasoning Using Stochastic Simulation of Causal models*. Technical Report R-68, CSD-8600##, Cognitive Systems Laboratory, UCLA Computer Science Department, September, 1986.

[21] Pearl, J. Fusion, propagation, and structuring in belief networks. *Artificial Intelligence* 29:241-288, 1986.

[22] Rousseau, W.F. *A method for computing probabilities in complex situations*. Technical Report 6252-2, Stanford University Center for Systems Research, May, 1968.

[23] Shachter, R.D. Intelligent probabilistic inference. In Kanal, L. and Lemmer, J. (editors), *Uncertainty in Artificial Intelligence*, . North Holland, 1986.

[24] Simon, H. A. A Behavioral Model of Rational Choice. *Quarterly Journal of Economics* 69:99-118, 1955.

[25] Smith, D.E. *Controlling inference*. Technical Report STAN-CS-86-1107, Stanford University, April, 1986.

[26] Szolovits, P. and Pauker, S.G. Categorical and probabilistic reasoning in medical diagnosis. *Artificial Intelligence* 11:115-144, 1978.

[27] Treitel, R. and Genesereth, M. R. Choosing Directions for Rules. In *Proceedings of the AAAI*. AAAI, Morgan Kaufman, Palo Alto, California, August, 1986.

[28] Tribus, M. *Rational Descriptions, Decisions and Designs*. Pergamon Press, New York, 1969.

[29] von Neumann, J., and Morgenstern, O. *Theory of Games and Economic Behavior*. Wiley, New York, 1953. (3rd edition).